\pdfoutput=1
\documentclass[11pt,dvipsnames]{article}

\usepackage[final]{acl}

\usepackage{times}
\usepackage{latexsym}

\usepackage[T1]{fontenc}
\usepackage[utf8]{inputenc}

\usepackage{microtype}

\usepackage{inconsolata}

\usepackage{graphicx}

\usepackage{hyperref}
\usepackage{makecell}
\usepackage[linesnumbered,ruled]{algorithm2e}
\usepackage{colortbl}
\usepackage{float}
\usepackage{stfloats}
\usepackage{enumitem}
\usepackage{booktabs}
\usepackage{multirow}
\usepackage{threeparttable}
\usepackage{amssymb}

\title{Know the Unknown: An Uncertainty-Sensitive Method for LLM
Instruction Tuning}

\author{
Jiaqi Li\textsuperscript{1},  
Yixuan Tang\textsuperscript{1},   
Yi Yang\textsuperscript{1}\\
\textsuperscript{1}The Hong Kong University of Science and Technology \\
}

\begin{document}
\maketitle
\begin{abstract}
Large language models (LLMs) demonstrate remarkable capabilities but face challenges from hallucinations, which typically arise from insufficient knowledge or context. 
While instructing LLMs to acknowledge knowledge limitations by responding with \emph{"I don't know"} appears promising, we find that models consistently struggle with admitting knowledge gaps. This challenge may originate from current instruction datasets that emphasise answer generation over knowledge boundary awareness. To address this limitation, we introduce \textbf{U}ncertainty-and-\textbf{S}ensitivity-Aware \textbf{Tuning} (\textbf{US-Tuning}), a novel two-stage approach for contextual question answering (QA). The first stage enhances LLMs' ability to recognise their knowledge boundaries, while the second stage reinforces instruction adherence through carefully designed causal prompts. Our experimental results demonstrate that US-Tuning not only significantly reduces incorrect answers in contextual QA but also improves models' faithfulness to their parametric knowledge, mitigating hallucinations in general QA tasks. Our fine-tuned Llama2-7B model achieves up to a 34.7\% improvement in handling out-of-knowledge questions and outperforms GPT-4 by 4.2\% in overall performance.
\footnote{The model and code are avaliable at \url{https://github.com/JiaqiLi404/TrustworthyRAG}}
% A promising solution involves instructing LLMs to respond with \emph{"I don't know"} when faced with questions beyond their knowledge scope. However, we observe that LLMs struggle to admit their lack of knowledge.

% primarily due to the design of existing instruction datasets, which prioritize generating specific answers over admitting uncertainty. 

% To address this, we propose a novel approach termed \textbf{U}ncertainty-and-\textbf{S}ensitivity-Aware \textbf{Tuning} (\textbf{US-Tuning}). 
% This method, tailored for contextual question answering (QA), involves a two-stage training aimed at inducing rejecting answering in models without compromising general QA capabilities. 
% The first stage focuses on enabling LLMs to recognize unknown questions, while the second stage reinforces their ability to adhere to instructions through carefully designed causal instructions. 
% Experimental results demonstrate that US-Tuning significantly reduces incorrect answers in contextual QA. Moreover, it improves the model's faithfulness to parametric knowledge, thereby mitigating hallucinations in general QA. Specifically, it achieves a substantial improvement of up to 34.7\% in handling unknown questions and even surpasses GPT-4 with an overall enhancement of up to 4.2\%.

\end{abstract}

\section{Introduction}
Large language models (LLMs) have demonstrated remarkable capabilities across a wide range of natural language processing tasks \cite{Brown_etal_Advances,wei2022finetuned,joshi-etal-2017-triviaqa}. Despite their impressive performance, these models face significant challenges that limit their reliable deployment in real-world applications. One of the most critical challenges is hallucination, the tendency to generate factually incorrect or non-sensical content \cite{maynez2020faithfulness}. This phenomenon occurs when LLMs generate outputs that either contradict the input context or introduce factually unsupported claims \cite{ji2023survey,ye2023cognitive}. The root cause of this behaviour lies in the inherent limitations in how these models learn and store knowledge during training. Specifically, LLMs encode extensive knowledge from training corpora, this knowledge is inherently incomplete and outdated. When encountering queries that require information beyond their knowledge, these models often resort to generating plausible but factually incorrect responses \cite{huang2023survey}.

% Large language models (LLMs) have shown great potential in various tasks \cite{Brown_etal_Advances,wei2022finetuned,joshi-etal-2017-triviaqa}. However, challenges such as hallucinations remain prevalent \cite{ji2023survey,ye2023cognitive}. Hallucination refers to the phenomenon in which LLMs generate factually incorrect or non-sensical responses \cite{maynez2020faithfulness}. This phenomenon is unavoidable, as LLMs acquire extensive knowledge from contexts or training corpora that inherently have limitations. When faced with unfamiliar information, LLMs tend to generate nonsense responses \cite{huang2023survey}.

\begin{figure}[t] 
\centering 
\includegraphics[width=0.495\textwidth]{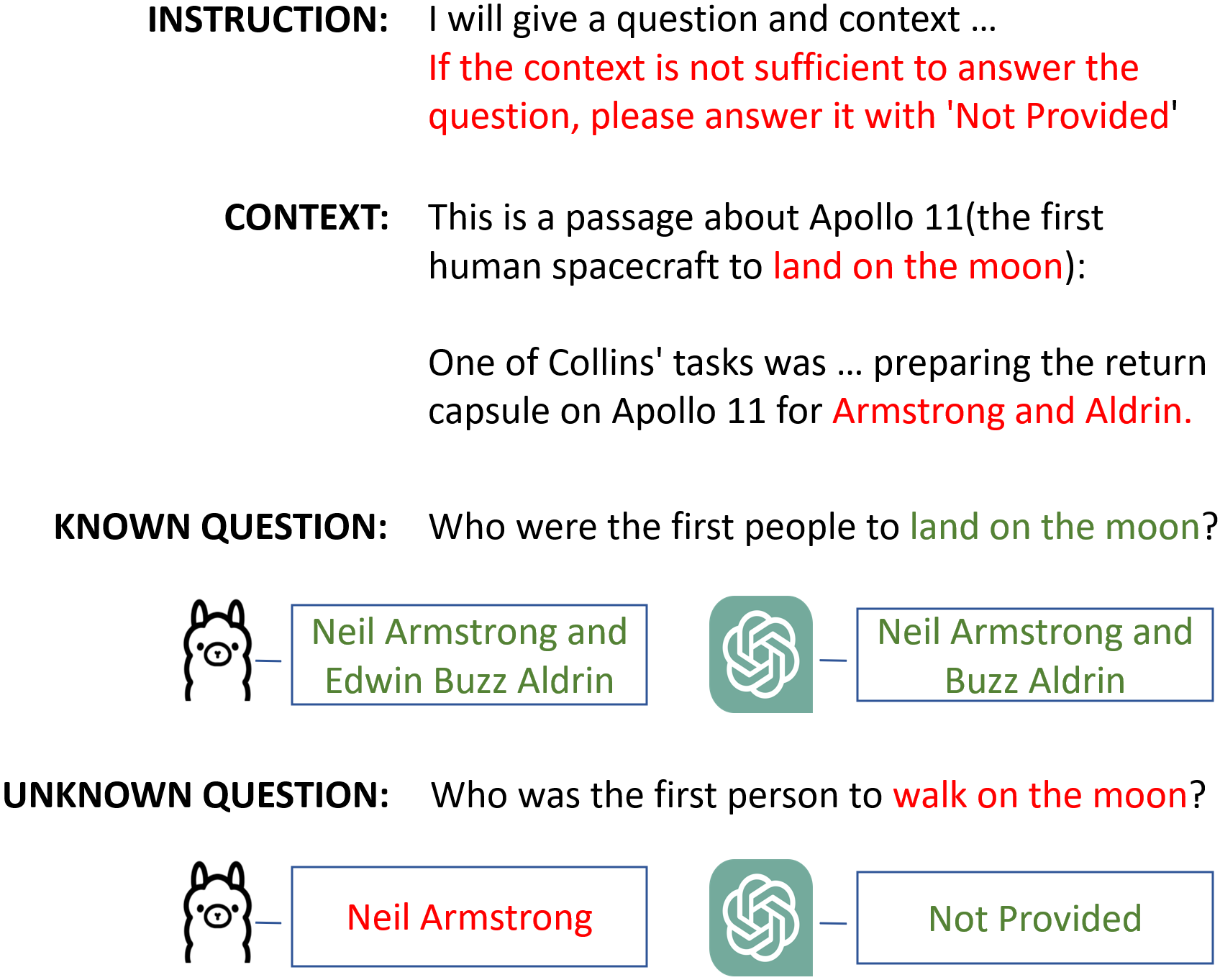} 
\caption{The intention of this paper is to address the inability of LLMs to recognise uncertain answers. We categorise questions into two types: \textbf{Known Questions}, which have specific answers, and \textbf{Unknown Questions}, which fall outside the provided context.} 
\label{Fig.intro-example} 
\end{figure}

To solve this question, two approaches have emerged. The first involves further fine-tuning models with additional knowledge \cite{liu2023webglm,gao-etal-2023-enabling,liu-etal-2023-evaluating}, while the second leverages retrieval-augmented generation techniques to incorporate external databases \cite{es2023ragas}. However, as demonstrated in Fig. \ref{Fig.intro-example}, these approaches still struggle with unknown queries in real-world applications, often producing incorrect answers. Recent work suggests that LLMs should be capable of acknowledging their knowledge limitations by explicitly stating "I don't know" when applicable \cite{cole-etal-2023-selectively,yu2024reeval}. However, there are two major challenges. First, current instruction datasets predominantly train LLMs to provide definitive answers, inadvertently discouraging models from recognising and expressing uncertainty—defined here as a model's awareness of knowledge beyond its training boundaries \cite{zhang2024rtuning}.
Second, models  explicitly optimised for uncertainty recognition often exhibit degraded performance in zero-shot question answering (QA) \cite{kasai2023realtime,li-etal-2023-large,si2023prompting}. A fundamental barrier to addressing these challenges is the lack of high-quality datasets containing unknown questions for training and evaluation. Thus, in this work, we focus on constructing contextual QA, including a scenario where the provided context is intentionally insufficient. We prioritise this approach over regulating parametric knowledge due to its greater impact on reasoning processes \cite{huang2024enhancing}. 

Our dataset development is motivated by research showing that subtle discrepancies between available knowledge and questions can trigger hallucinations \cite{shuster-etal-2021-retrieval-augmentation}. Building on the ASQA data set \cite{stelmakh-etal-2022-asqa}, we create a balanced collection of both in-context (known) and out-of-context (unknown) questions. For the latter, we deliberately introduce minor inconsistencies in the context, such as mismatched dates or objects, while maintaining overall contextual coherence. Unlike previous works \cite{li2022large,chen2023purr}, these subtle discrepancies are particularly effective in exposing the tendency of LLMs to hallucinate, making our data set especially valuable for evaluating model performance.

To enhance LLMs' capability to know the unknown and reject uncertain answers, we introduce a novel training framework termed Uncertainty-and-Sensitivity-Aware Tuning (US-Tuning). This approach contains a two-stage training process designed to balance the trade-off between uncertainty recognition and zero-shot instruction adherence. By doing so, it enhances the ability to identify and acknowledge uncertainty while preserving its original QA performance. In the first stage, we focus on awareness of uncertainty, guiding LLMs to effectively identify questions outside the knowledge boundaries. The second stage emphasisess the sensitivity of the instruction, teaching the model to reject answering unknown questions and restoring the compromised QA performance through additional fine-tuning.

Our approach addresses several fundamental challenges in developing uncertainty-aware language models for question-answering tasks. The primary challenge lies in the delicate balance between admitting the knowledge boundary and general QA performance—models that are overly sensitive to uncertainty often experience significant degradation in their ability to answer standard questions. Additionally, when fine-tuning uncertainty-aware models on conventional QA datasets, which contain questions with supporting evidence, models frequently lose their ability to effectively recognise and reject unknown queries. We attribute this degradation to the model's weak sensitivity to uncertain instructions and address it through carefully designed causal instructions in our approach.

% Our approach addresses several key challenges.
% First, models that are overly sensitive to uncertainty may experience degradation in general QA performance.
% Second, fine-tuning uncertainty-aware models on standard QA datasets, which primarily consist of known queries, often leads to a significant decline in their ability to reject answering unknown questions.
% We attribute this degradation to the model's weak sensitivity to uncertain instructions and incorporate carefully designed causal instructions in the second stage to address it.
Experimental results demonstrate that US-Tuning significantly improves the performance of prevalent LLMs in acknowledging the unknown. Notably, it achieves a 34.7\% improvement in addressing unknown questions and surpasses GPT-4 \cite{OpenAI} with an overall performance increase of up to 4.2\%. Furthermore, it not only reduces the frequency of incorrect answers in contextual QA but also encourages LLMs to remain faithful to their parametric knowledge, thereby mitigating hallucinations across various benchmark assessments. Our key contributions are as follows:
\begin{itemize}[itemsep=0pt,topsep=0pt,parsep=0pt,leftmargin=10pt]
\item We construct a novel dataset and benchmark for uncertainty recognition, enabling the evaluation of the models' awareness of knowledge gaps.
\item We investigate why LLMs tuned to prioritise uncertainty fail to adhere to essential instructions, attributing this behaviour to their weak sensitivity to uncertain prompts.
\item We propose a novel two-stage fine-tuning paradigm for instructing the model to remain faithful to the context and reject unknown questions while exploring the relationship between faithfulness and hallucinations.
\end{itemize}

% The primary objective of US-Tuning is to enhance the model's ability to identify and acknowledge uncertainty while maintaining the original QA performance.

\section{Related Work}
In this section, we analyse the former works about hallucinations and instruction datasets for training.

\subsection{Uncertainty in Hallucinations}
Although the large language models (LLMs) have demonstrated strong performance in downstream tasks by generalising and leveraging encoded knowledge within the parameters \cite{pu-demberg-2023-chatgpt,zhang2023hallucination}, 
the uncertainty of such knowledge can also mislead models to generate untrustworthy outputs \cite{yu2024improving,ye2023cognitive,manakul-etal-2023-selfcheckgpt}.
Generally, the uncertainty comes from training data and overestimation \cite{zhang2024rtuning}.
Research shows that models tend to mimic the output in the training set \cite{kang-hashimoto-2020-improved}, leading to hallucinations that generate reasonable answers for insufficient question-context pairs.
Furthermore, models could be overconfident in their capacities and fail to identify unknown questions \cite{yin2023large,ren2023investigating,kadavath2022language}.

There are studies focusing on uncertainty measurement to mitigate hallucinations.
\citet{lu2023prompts} conclude that a correlation exists between the uncertainty and the accuracy.
CAD \cite{shi2023trusting} proposes a contrastive method for measuring the uncertainty of generated knowledge, restricting models to be context-awarded by amplifying output probabilities when the context is provided.
SelfCheckGPT \cite{manakul-etal-2023-selfcheckgpt} utilises sampling to identify and exclude uncertain information.

\subsection{Faithfulness to the External Knowledge}
Hallucination is defined as generations that are nonsensical or unfaithful to the provided source content \cite{ji2023survey,filippova2020controlled}, encompassing both context and paremetic knowledge.
While most prior research has concentrated on the model's faithfulness to parametric knowledge, the aspect of contextual faithfulness as a specific and significant form of hallucination has received comparatively less attention.
This gap is underscored by findings indicating that the incorporation of up-to-date and relevant knowledge within prompts can effectively mitigate fact-conflicting hallucinations \cite{zhou-etal-2023-context,liu2022tokenlevel}.
However, these studies \cite{vu2023freshllms,lewis2020rag} operate under the assumption that the given context is always sufficient for generating accurate answers. To address this limitation, various approaches utilise LLMs for post-generation detection \cite{shen2023news} or editing \cite{chen2023purr} to ensure the faithfulness and consistency of the generated responses with the provided contexts.
Self-RAG \cite{asai2023selfrag} leverages LLMs to screen the provided context, avoiding the disruptions of irrelevant information. However, models struggle to accurately determine whether the provided knowledge is sufficient for answering, especially when the domains of query and context exhibit similarities. 
Furthermore, some research suggests that reliance on 'unknown' external knowledge can significantly impair performance, potentially exacerbating hallucinations \cite{lee2024crafting}. Thus, there is a pressing need for an LLM capable of knowing the 'unknown'.

\subsection{Instruction Dataset for Training}

Aligning LLMs necessitates substantial training data, prompting a trend toward synthesising instruction data to enhance performance. Self-Instruct \cite{wang2023selfinstruct} proposes generating diverse instructions using ChatGPT. To improve the query's complexity in different dimensions, WizardLM \cite{xu2023wizardlm} uses five prompts, including depth search and with search. Conversely, AttrPrompt \cite{yu2023large} generates various instructions from a feature perspective without relying on class-conditional prompts. Most existing methods concentrate on improving answer quality by exploring a variety of questions with definitive answers, rather than addressing where answers are uncertain. Recent research \cite{zhang2024rtuning, cole-etal-2023-selectively} has led LLMs to reject unknown questions. R-Tuning \cite{zhang2024rtuning}, for example, trains models to recognise their knowledge limits and to respond with "I don’t know".
However, identifying the boundaries of parametric knowledge remains challenging due to factors such as latent space compression and hallucination. Therefore, in this study, we build a dataset based on contextual question answering and propose a two-step training method that enables models to reject unknown questions while preserving performance in other tasks.

\section{Uncertainty-and-Sensitivity-Aware Tuning}
\label{sec:methodology}
% This section focuses on our proposed Uncertainty-Aware Tuning (UT) and Sensitivity-Aware Tuning (ST). 
% First, we instruct models to assess the boundary of knowledge. Next, we use prompt-sensitive tuning to improve the model's ability to follow specific instructions. By integrating these two stages, we effectively train large language models (LLMs) to acknowledge their limitations without compromising their question-answering abilities. 

\begin{figure*}[ht]
  \centering  \includegraphics[width=0.99\textwidth]{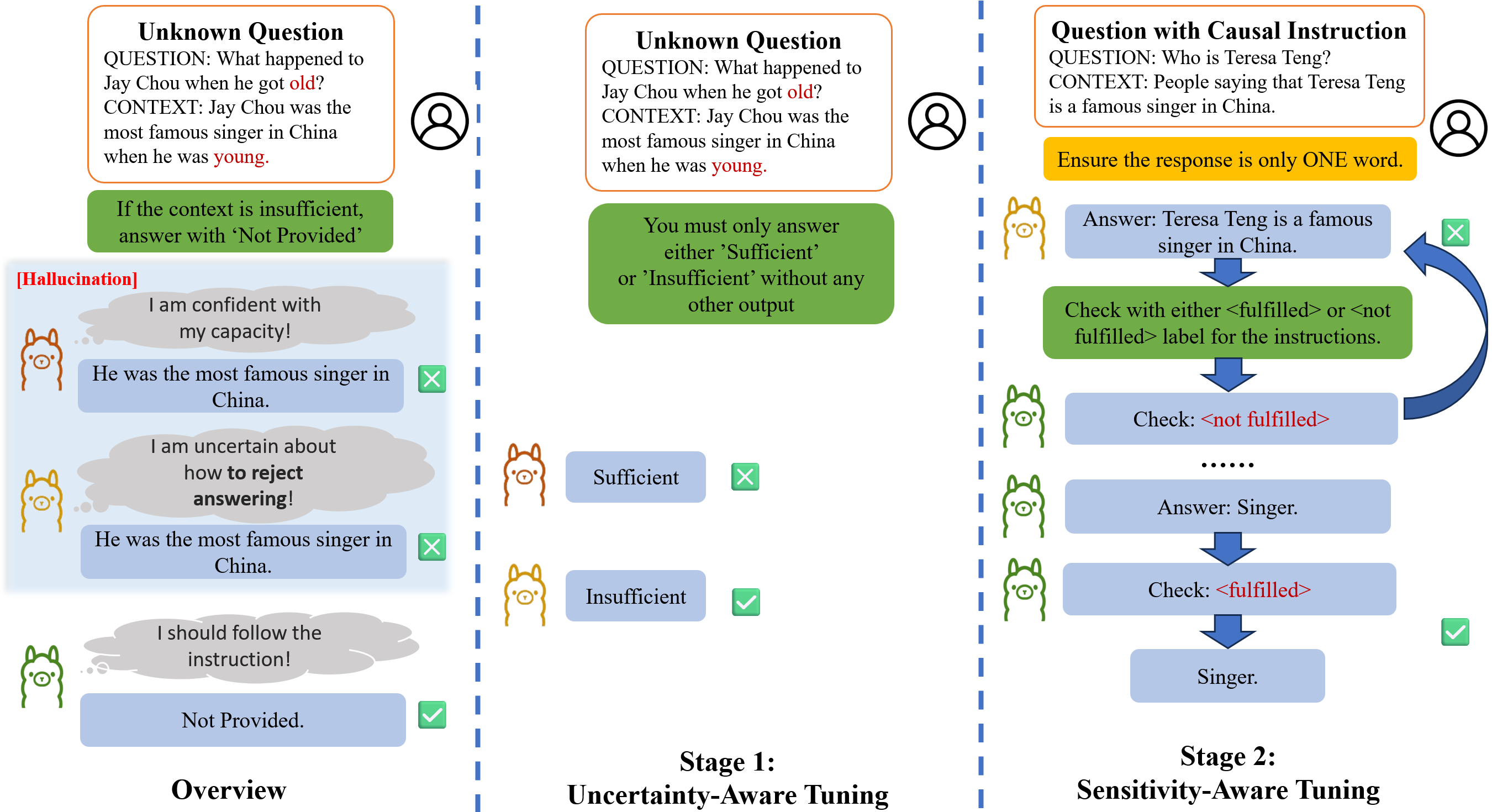}
  \caption{Illustration of our US-Tuning. The \colorbox{YellowGreen}{green dialog boxes} represent task-oriented instructions, while the \colorbox{Dandelion}{yellow box} indicates additional causal instructions influencing the output. 
  \textbf{Overview:} The models include the \textcolor{Bittersweet}{vanilla model}, the \textcolor{Dandelion}{Uncertainty-Aware Tuned (UT) model}, and the \textcolor{ForestGreen}{Sensitivity-Aware Tuned (ST) model}. We highlight that hallucinations stem from weak cognition of uncertainty and ignorance of instructions.
  \textbf{UT (Stage 1):} teaching the model to know the unknown. 
  \textbf{ST (Stage 2):} instructing the model to effectively follow provided instructions. }
  \label{fig:PSQA}
\end{figure*}

Our research centres on the open-book contextual question-answering (QA), which aims to generate an answer $a$ based on three inputs: $i_{t}$, $q$, and $c$. Here, $i_{t}$ denotes the task instructions, $q$ represents the question, and $c$ refers to the provided context. The generation process $G$ can be formulated as: 
$$a = G(i_{t}, q, c)$$

To induce the model to analyse uncertainty, we will implement two explicit constraints. First, we instruct the model not to utilise knowledge beyond the context by stating in $i_{task}$: \emph{"Your answer must not use any additional knowledge that is not mentioned in the given contexts".} Second, we require the model to reject uncertain answers with the directive: \textit{"If the context is not sufficient to answer the question, please answer it with ’Not Provided’"}. This process relies on the model $G$ to evaluate whether the context $c$ is adequate to answer the question $q$. Based on this assessment, $G$ either generate an appropriate response ($a$) or acknowledge the insufficiency of $c$.

\subsection{Motivation}

As demonstrated in Table \ref{table:benchmark}, our benchmark indicates that vanilla large language models (LLMs) exhibit limited efficacy in rejecting questions beyond their knowledge boundaries. Through systematic experimentation, we identify two core challenges underlying this limitation. First, models frequently generate speculative answers to satisfy perceived user expectations, attributable to standard QA training paradigms that prioritise definitive responses over uncertainty acknowledgement. 
Second, models fine-tuned for uncertainty recognition demonstrate weakened adherence to the zero-shot instructions, creating a trade-off between rejecting unknown questions and generalisable instruction-following capabilities.
This trade-off arises from the scarcity of highly confusing unknown question-context pairs. To preserve the integrity of these rare but critical samples, we avoid direct fine-tuning on unknown questions. 
Instead, our proposed two-stage training framework addresses these challenges synergistically. The first stage emphasises training the model to identify and reject uncertain questions, thereby preventing inaccurate responses. The second stage involves a systematic instruction review process with answer refinement, contrasting conventional QA tuning by emphasising instruction adherence in response generation.
%Figure \ref{fig:PSQA} shows the overview of the proposed two-stage method.

% \begin{table}[h]\LARGE
% \renewcommand\arraystretch{1.5}
% \resizebox{\columnwidth}{!}{%
% \begin{tabular}{llcc}
% \toprule[2pt]
% % \hline
% \multicolumn{1}{c}{Model}           & $Acc_{known}$ & $Acc_{unknown}$ & $F1$ \\ \hline
% Baseline                            & \textbf{79.3}         & 58.3           & \textbf{67.2} \\ \hline
% Balance Dataset Finetuned           & 52.4         & \textbf{84.4}           & 64.6 \\ \hline
% \makecell[l]{Balance Dataset Finetuned \\+ HotpotQA} & 77.0         & 20.9           & 32.8 \\ 
% % \hline
% \bottomrule[2pt]
% \end{tabular}%
% }
% \caption{Results (in \%) for the preliminary experiment. The baseline is the vanilla Llama2-7B-Chat. We report the accuracy of different questions in the benchmark (detailed in section \ref{sec:experiments}). }
% \label{tab:exploration}
% \end{table}

\subsection{Stage 1: Uncertainty-Aware Tuning (UT)}
\label{section:ur}

The first stage fine-tunes the model to accurately recognise its knowledge boundaries and identify the known questions. To safeguard the ground truth in the benchmark, we formalise this task as a binary classification problem, as shown in Figure \ref{fig:PSQA}. Questions are categorised into two groups: known questions and unknown questions. Known questions are defined as queries with sufficient contextual support to yield accurate answers. Conversely, unknown questions are characterised by lacking adequate contextual information, often exhibiting subtle differences from the query. The model learns to evaluate contextual adequacy and classify its confidence as either "Sufficient" or "Insufficient" for response generation.

Formally, given a contextual QA dataset $D=\{(q_{i},c_{i}),(q_{i},{c^{\prime}}_{i})\}_{i=1}^n$ comprising $n$ known question-context pairs and $n$ unknown pairs, we fine-tune the LLM to perform binary classification, where responses are restricted to two categories: "Sufficient" and "Insufficient." 
The instruction for tuning is recorded in Appx. \ref{sec:appendix_instructions:CA}.

\subsection{Stage 2: Sensitivity-Aware Tuning (ST)}
\label{section:ps}

Although UT enables models to delineate knowledge boundaries and reject unanswerable queries, Table \ref{table:benchmark} reveals two critical challenges.
First, UT-trained models exhibit heightened uncertainty sensitivity, which affects their ability to answer known questions with confidence.
Second, conventional QA tuning exacerbates the model’s inability to reject unknown questions, as UT reduces sensitivity to uncertain instructions. 
We hypothesise that this stems from a conflict in objective alignment: instructions for rejecting unknowns (applicable only to out-of-distribution queries) are not effective on the training data. Consequently, enforcing these instructions during evaluation introduces a misalignment between uncertainty recognition and instruction adherence, degrading overall performance.

To address this, our proposed ST is motivated by explicitly distinguishing the instructions into causal and non-causal ones. 
\begin{itemize}
    \item \textbf{Causal} instructions directly affect the response content, whereas non-causal instructions provide auxiliary guidance without affecting answer semantics. For example, instructions that constrain the format or tense of responses serve as typical causal ones. Conversely, extra instructions, such as \emph{"answering with ’Not Provided’ if the context is insufficient"}, function as non-causal instructions when fine-tuning known questions, as they do not contribute directly to the answer. 
    \item \textbf{Non-causal }instructions risk being disregarded, despite their critical importance to the overall task. 
\end{itemize}

% Causal instructions directly affect the response content, whereas non-causal instructions provide auxiliary guidance without affecting answer semantics. For example, instructions that constrain the format or tense of responses serve as typical causal ones. Conversely, extra instructions, such as \emph{"answering with ’Not Provided’ if the context is insufficient"}, function as non-causal instructions when fine-tuning on known questions, as they do not contribute directly to the answer. Consequently, non-causal instructions risk being disregarded, despite their critical importance to the overall task. 
% This phenomenon parallels readers who perceive a book's introduction as irrelevant, leading them to underestimate its significance and neglect reading it, despite its crucial role in some specific books.
Our ST is designed to enhance the model's sensitivity and adherence to all instructions by ensuring that even non-causal instructions are prioritised. 
As shown in Fig. \ref{fig:PSQA}, it comprises two synergistic components: additional causal instructions and instruction review synthesis.

\textbf{Causal Instruction Synthesis:} By instructing GPT-4 to produce controlling conditions that directly influence response properties, such as tense, length, or output format, we obtain additional causal instructions. These causal instructions are then randomly integrated into the original QA prompts, ensuring the model learns to prioritise and comply with diverse task requirements.
The prompt for generation is presented in Appx. \ref{sec:appendix_instructions:PSGeneration}.

\textbf{Review Instruction Synthesis:} The instruction review module employs the model itself to verify the fulfilment of all instructions.
The model will recursively regenerate until it gets a perfect answer by utilising the prompts in Appx. \ref{sec:appendix_instructions:PS}.
The process of the instruction review is illustrated in Algo. \ref{algo:revise}.

As shown in {Fig. \ref{fig:PSQA}, given a question-answering dataset $\{(q_{1},c_{1}), ...(q_{n},c_{n})\}$ and additional causal instructions,
the entire process is formulated as $a=R(G(i_{t}+i_{c},q,c))$, where $i_{c}$ is a randomly selected casual instruction and $i_{t}$ is the original task description.
$R$ is the loop function for instruction review. We employ GPT-4 and record the conversation from the loop to fine-tune the smaller model.

\section{Experiments}
\label{sec:experiments}
In this section, we describe the data construction and the associated experiments. 
Table \ref{table:benchmark} shows that the suboptimal performance of LLMs in rejecting unknown questions can be attributed to two primary factors: weak uncertainty-recognition capacity and the instruction-sensitivity reduction.
We assess the effectiveness of US-Tuning using prevalent LLMs on our proposed benchmark, as well as on traditional QA hallucination benchmarks.

\subsection{Data Construction}
\label{sec:data_construction}
% We synthesise a benchmark balanced in terms of known and unknown questions for evaluation, along with two distinct datasets specifically designed for US-Tuning. 
We create a benchmark that balances known and unknown questions for evaluation, along with two specific datasets designed for US-Tuning.

\textbf{Uncertainty-Recognition Benchmark} To comprehensively evaluate the model's cognitive ability to identify knowledge gaps, we construct a test dataset using the ASQA \cite{stelmakh-etal-2022-asqa} dataset, which consists of ambiguous questions. Each question is divided into multiple sub-questions with their corresponding contexts. For example, as recorded in Appx. \ref{sec:appendix_dataset}, one pair may discuss the discovery of the photoelectric effect in 1887, while another may cover the theoretical development in 1905. To generate the unknown questions, we shuffle these pairs, reassigning the questions to different but related contexts. As a result, there are two significant advancements in our benchmark. First, the context is closely relevant to the query, featuring partial mismatches in dates or objects, thereby challenging the model's ability to handle uncertainty. Second, the context is definitely insufficient for the query. Such samples are rare and valuable, as ASQA is the only dataset we have found that could yield sufficient samples that satisfy the requirement. We generate 3,320 known questions and 3,320 unknown questions to construct our benchmark.

In the evaluation, we design the QA template for uncertainty recognition by instructing the model to reject unknown questions, as presented:

\begin{itemize}
\item QA Uncertainty-Recognition: \emph{If the context is not sufficient to answer the question, please answer it with ’Not Provided’.}
\end{itemize}

\textbf{US-Tuning Datasets} Two distinct instruction datasets are used for separate stages. For the UT, we construct a binary dataset comprising 646 samples from the ASQA \cite{stelmakh-etal-2022-asqa} with the ground truth concealed to prevent overlap with the evaluation data. Here is a demonstration of the prompt we used for tuning on this dataset:
\begin{itemize}
\item Uncertainty-Aware Tuning: \emph{You must only answer either ’Sufficient’ or ’Insufficient’ without any other output}
\end{itemize}

To protect our valuable benchmark, the dataset for ST is derived from HotpotQA \cite{yang2018hotpotqa}, a dataset designed for multi-hop QA. We generate causal instructions using GPT-4 \cite{OpenAI} and manually select the 28 most robust instructions, as listed in Appx. \ref{appx:causal_ins}. These instructions were then integrated into 300 randomly selected samples from HotpotQA. Subsequently, we utilised GPT-4, following the methodology outlined in Section \ref{section:ps}, to synthesise the final ST dataset.

\subsection{Experiment Setting}

\textbf{Training Details.} We evaluate our US-Tuning on prevalent open-sourced LLMs, including Llama2-7B-Chat \cite{touvron2023llama}, Mistral-7B-Instruct-v0.2 \cite{jiang2023mistral}, and Gemma-2-9B-Instruct \cite{team2024gemma}. 
We also test GPT-4o \cite{gpt4o}, GPT-4-1106-preview \cite{OpenAI}, GPT-3.5 Turbo \cite{gpt3.5}, Vicuna-7B v1.5 \cite{zheng2024judging} and Self-RAG-7B \cite{asai2023selfrag} on our benchmark.
Furthermore, we conduct experiments on different model sizes based on the latest Llama3-it \cite{llama3} and Gemma3-it \cite{team2025gemma} models.
Our fine-tuning bases on an RTX3090 GPU in conjunction with LLaMA-Factory \cite{zheng2024llamafactory}, with Lora \cite{hu2021lora} in a rank of 8, a batch size of 4, and a learning rate of 5e-5. We configured the epochs to 1 and 5 for the two stages, respectively.
This research integrates the instruction-based and attributed prompts, which demonstrate to effectively mitigate hallucinations \cite{zhou-etal-2023-context}, as provided in Appx. \ref{sec:appendix_instructions}.

\textbf{Evaluation Metric.} We use $Acc_{known}$ for representing the accuracy of questions with specific answers, and $Acc_{unknown}$ for unknown questions.

\textbf{Benchmark Result.} As summarised in Table \ref{table:benchmark}, our analysis (Appx. \ref{sec:results}) reveals that prevalent LLMs struggle to reliably identify unknown questions, achieving modest accuracy rates of ~60\%.

\subsection{Analysis}
\subsubsection{Weak Uncertainty-Recognition Capacity}

Tables \ref{table:benchmark} and \ref{table:binaryUR} reveal a persistent performance gap of up to 21.0\% between known and unknown questions for Llama2, indicating the challenge associated with models' capacity to recognise uncertainty. By leveraging uncertainty-aware tuning (UT), as evidenced in Table \ref{table:benchmark}, there is a notable improvement of up to 26.1\% in the accuracy of responses to unknown questions ($Acc_{unknown}$), surpassing baseline performances and being comparable to GPT-4. However, this increased awareness of uncertainty leads to a decrease in the QA capability. 
Specifically, models demonstrate an excessive sensitivity to the varied phrasing of similar questions.

\subsubsection{Instruction-Sensitivity Reduction Problem}
\label{sec:instruction-sensitivity-reduction-problem}

According to Table \ref{table:benchmark}, further fine-tuning on HotpotQA results in a degradation in the model’s ability to reject unknown questions, primarily due to a decline in its adherence to instructions. This is evidenced by a low $Acc_{unknown}$ of 20.9\%, despite the uncertainty recognition capacity being maintained at 66.7\% (Table \ref{table:binaryUR}). 
We term this phenomenon the "instruction-sensitivity reduction problem."

As shown in Tables \ref{table:benchmark} and \ref{table:binaryUR}, UT equips the model with the ability to recognise and reject uncertain questions. However, the absence of unknown questions in HotpotQA means that the instruction to reject uncertain answers is never effectively implemented during training. This creates a conflict that adherencing to zero-shot instructions can inadvertently increase uncertainty, counteracting the objectives of UT and diminishing performance. Consequently, the model often disregards instruction constraints, generating hallucinated answers for unknown questions. Our proposed ST (US-Tuning in Table \ref{table:benchmark}) addresses this issue by ensuring adherence to all instructions, bridging the gap between uncertainty recognition and instruction compliance.

\begin{table}[t]
\centering
\resizebox{0.50\textwidth}{!}{\begin{tabular}{llccc}
\hline
& & \multicolumn{3}{c}{QA Uncertainty-Recognition}\\
Category                    & Model                & $Acc_{known}$ & $Acc_{unknown}$  & $F1$ \\
\hline
\multirow{5}{*}{Benchmark}  & \textbf{GPT-4o} & 80.2 & \textbf{85.6} & \textbf{82.8} \\
                            & GPT-4 & 79.6 & 83.6 & 81.6 \\
                            & GPT-3.5       & \textbf{82.1} & 51.8 & 63.5 \\
                            & Vicuna-7B v1.5       & 74.6 & 43.8 & 55.2 \\
                            & Self-RAG-7B          & 67.9 & 48.1 & 56.3 \\
\hline
\multirow{4}{*}{Llama2}     & Vanilla   & 79.3 & 58.3 & 67.2 \\
                            & UT (Stage 1)         & \cellcolor{orange!20}52.4 & \cellcolor{green!20}84.4 & 64.6 \\
                            & UT+HotpotQA        & 77.0 & \cellcolor{orange!20}20.9 & \cellcolor{orange!20}32.8 \\
                            & \textbf{US-Tuning}   & \textbf{79.7} & \cellcolor{green!20}\textbf{93.0} & \cellcolor{green!20}\textbf{85.8}\\
\hline
\multirow{4}{*}{Mistral}    & Vanilla   & 85.1 & 63.0 & 72.4\\
                            & UT (Stage 1)         & \cellcolor{orange!20}77.5 & \cellcolor{green!20}\textbf{75.8} & 76.6\\
                            & UT+HotpotQA        & 87.1 & \cellcolor{orange!20}52.4 & \cellcolor{orange!20}65.5 \\
                            & \textbf{US-Tuning}   & \textbf{87.3} & \cellcolor{green!20}75.3 & \cellcolor{green!20}\textbf{80.9} \\
\hline
\multirow{4}{*}{Gemma}      & Vanilla   & 86.1 & 74.1 & 73.5\\
                            & UT (Stage 1)         & \cellcolor{orange!20}76.1 & \cellcolor{green!20}\textbf{86.2} & \cellcolor{green!20}80.8 \\
                            & UT+HotpotQA        &\cellcolor{green!20} \textbf{91.3} & \cellcolor{orange!20}20.8 & \cellcolor{orange!20}33.9 \\
                            & \textbf{US-Tuning}   & 87.6 & \cellcolor{green!20}81.2 & \cellcolor{green!20}\textbf{84.3} \\
\hline
\end{tabular}}
\caption{Results (in \%) for prevalent LLMs on QA uncertainty-recognition benchmark. The overall best results for each category are highlighted in \textbf{bold}.
Results that are more than 5\% higher or lower than the baseline are highlighted in \colorbox{green!20}{green} and \colorbox{orange!20}{orange}, respectively.}
\label{table:benchmark}
\end{table}

\begin{table}[t]
\large
\centering
\resizebox{0.50\textwidth}{!}{
\begin{threeparttable}
\begin{tabular}{llcccccc}
\hline
 & & \multicolumn{3}{c}{Vanilla}& \multicolumn{3}{c}{\textbf{US-Tuning (ours)}} \\
Model & Size & Kno. & Unk.  & $F1$ & Kno. & Unk.  & $F1$ \\
\hline
\multirow{3}{*}{Gemma3} & 1B & 81.6 & 2.8 & 5.4 & 81.8 & \textbf{40.3} & \textbf{54.0} \\
& 4B & 88.0 & 38.7 & 53.8 & \textbf{82.3} & \textbf{68.6} & \textbf{74.9} \\
& 12B & 94.2 & 57.4 & 71.3 & \textbf{84.7} & \textbf{79.9} & \textbf{82.3} \\
\multirow{3}{*}{Llama3} & 1B & 74.9 & 4.6 & 8.6 & \textbf{81.4} & \textbf{17.0} & \textbf{28.0} \\
& 3B & 86.4 & 51.4 & 64.5 & 84.3 & \textbf{63.8} & \textbf{72.7} \\

 & 8B\tnote{*} & 82.8 & 74.6 & 78.5 & 86.6 & 78.6 & 82.4 \\
\hline
\end{tabular}
\begin{tablenotes}
        \footnotesize
        \item * The 8B model is based on Llama 3.1, whereas the remaining models are fine-tuned on Llama 3.2
      \end{tablenotes}
  \end{threeparttable}
  }
\caption{Accuracies (in \%) on known and unknown questions for the LLMs of different sizes on QA uncertainty recognition benchmark. Values that are more than 5\% better than vanilla model are highlighted in \textbf{bold}.}
\label{table:comp:size}
\end{table}

\begin{table}[b]
\large
\centering
\resizebox{0.45\textwidth}{!}{\begin{tabular}{lcccccc}
\hline
 & \multicolumn{3}{c}{Vanilla} & \multicolumn{3}{c}{\textbf{US-Tuning (ours)}}\\
Model & Cor. & Wro. & Unk. & Cor. & Wro. & Unk. \\
\hline
GPT-4  & 79.6 & 4.4 & 16.0& - & - & - \\
\hline
Llama2  & 79.2 & 8.5 & 12.2& 79.7 & \textbf{1.4} & 18.9  \\
Mistral  & 85.1 & 5.1 & 9.8 & 87.3 & \textbf{2.5} & 10.2 \\
Gemma & 86.1 & 3.9 & 10.0& 87.6 & \textbf{1.6} & 10.8  \\
\hline
\end{tabular}}
\caption{The portions of correct, wrong, and unknown responses among the responses for known questions.}
\label{table:comp:pos}
\end{table}

\begin{table}[t]
\centering
\resizebox{0.50\textwidth}{!}{\begin{tabular}{llcccc}
\hline
Category & Method & Samples & Kno. & Unk. & F1 \\
\hline
Vanilla                     & Llama2  &    -& 79.3 & 58.3 & 67.2 \\
\hline
\multirow{2}{*}{Post-Gen.}  & Validation&    -  & 82.5 & 53.8 & 65.1 \\
                            & Sampling&    -  & 79.7 & 66.5 & 72.5 \\
\hline
Prompt                      & CFP&    -  & \textbf{87.4} & 47.6 & 61.6 \\
\hline
\multirow{4}{*}{Tuning}     & Calibration &    -       & 67.1    & 63.2          & 65.1       \\
                            & Honesty  &    8k          & 74.7    & 61.1          & 67.2        \\
                                    
                            & C-DPO    &    4k          & 77.7    & 69.6          & 73.4         \\
                            & \textbf{US-Tuning} &    3k & 79.7    & \textbf{93.0} & \textbf{85.8}\\
\hline
\end{tabular}}
\caption{Accuracies (in \%) on known and unknown subsets for SOTA methods on the QA uncertainty recognition benchmark. We also report the cumulative number of training samples used by each SOTA method.}
\label{table:comp:post-generation}
\end{table}

\subsection{Effectiveness on Contextual QA}

Among the models tested on our benchmark, the US-Tuned Llama2 ranks the highest, achieving an F1 score of 85.8\%, which surpasses GPT-4 by 4.2\% and exceeds the baseline by 18.6\% (as shown in Table \ref{table:benchmark}). This impressive performance can be attributed to the model's optimal balance between uncertainty recognition and adherence to zero-shot instructions. Notably, it achieves a remarkable 93.0\% accuracy on unknown questions, the highest among prevalent LLMs, while maintaining a 79.7\% accuracy on known questions.
Additionally, results in Table \ref{table:comp:size} show that our method is robust and generalises well across various models and sizes, especially for Gemma3, which receives significant performance gain, highlighting the robustness and effectiveness of our US-Tuning approach in enhancing performance among prevalent LLMs. This tuning method effectively mitigates the risk of generating incorrect answers without compromising the original question-answering capabilities.

Our model effectively supports high-stakes decision-making. For unknown questions, in addition to the significantly increased $Acc_{unknown}$, the case study in Appx. \ref{sec:casestudy} demonstrates that our model prioritises uncertainty analysis, acknowledging limitations rather than hallucinating responses. For known questions, Table \ref{table:comp:pos} presents a detailed distribution of responses. The data indicate that US-Tuning substantially reduces the occurrence of wrong answers by up to 7.1\%, albeit with a modest increase in the proportion of unknown responses.

\subsection{Comparison with SOTA Approaches}
\label{sec:exp:sotacomperision}

We evaluate our method against SOTA approaches within our uncertainty-recognition benchmark, as detailed in Appx. \ref{sec:sotamethods}. Table \ref{table:comp:post-generation} shows that Honesty \cite{yang2023alignment} and Calibration \cite{kapoor2024calibration}, which target noncontextual QA tasks, face significant instruction-sensitivity reduction, evidenced by the low $Acc_{unknown}$. Despite being fine-tuned with unknown questions, these methods prioritise uncertainty but struggle with uncertain zero-shot instructions related to contextual uncertainty identification. As a result, they exhibit limited robustness in contextual QA. 
However, when integrated with our proposed ST, as experimented in Appx. \ref{sec:st_honesty}, Honesty exhibits significantly improved compliance with instructions and outperforms the baseline.
This highlights the effectiveness of our ST in generalising uncertainty recognition capacity across diverse tasks.
The results of C-DPO \cite{bi2024context} indicate that Direct Preference Optimisation \cite{rafailov2024direct} effectively enhances the overall capabilities of the model in both QA and instruction adherence, but a gap persists compared to our tailored method.
Additionally, post-generation methods face challenges in recognising unknown questions due to their limited capacity for uncertainty detection.

We also report the cumulative number of training samples used by each method in Table \ref{table:comp:post-generation}. The cumulative number is computed as the product of the number of training samples and the number of training epochs. For example, in our US-Tuning approach, 1,292 samples are used for one epoch in UT, and 300 samples are used for five epochs in ST, resulting in a cumulative total of 2,792 samples.
The results demonstrate that our approach is less time-consuming, showing that small but high-quality datasets can yield effective performance.

\subsection{Ablation Study}
\label{sec:ablation}
To further investigate the impact of US-Tuning, we decompose it into three distinct components.
The detailed explanations are recorded in Appx. \ref{appx:ablation}.

\begin{itemize}[itemsep=0pt,topsep=0pt,parsep=0pt,leftmargin=10pt]
\item \textbf{UT}: 646 samples for uncertainty-aware tuning.
\item \textbf{HP}: 300 samples from HotpotQA with QA prompts provided in Appx. \ref{sec:appendix_instructions:QA}.
\item \textbf{CI}: HP with causal instructions, termed ST.
\end{itemize}

As illustrated in Table \ref{table:ablation} and Fig. \ref{fig:ablation}, our findings indicate that models without UT exhibit a weak capacity for uncertainty recognition, presented by low $Acc_{unknown}$. 
Furthermore, QA fine-tuning that does not incorporate causal instructions contradicts the objectives of UT, resulting in a decline in $Acc_{unknown}$. 
In contrast, our ST approach not only enhances performance on known answers, achieving the highest $Acc_{known}$ reported in the table. But also, when effectively integrated with UT, our method attains optimal performance across both known and unknown questions.

\begin{table}[H]
\large
\centering
\resizebox{0.41\textwidth}{!}{\begin{tabular}{ccc|ccc}
\hline
\multicolumn{3}{c|}{Component} & \multicolumn{3}{c}{QA Uncertainty-Recognition}\\
     UT     &     HP     &     CI     & $Acc_{known}$ & $Acc_{unknown}$ & $F1$ \\
\hline
            &            &            & 79.3 & 58.3 & 67.2 \\
\hline
 \checkmark &            &            & \cellcolor{orange!20}52.4 & \cellcolor{green!20}84.4 & 64.6 \\
            & \checkmark &            & 77.5 & 58.3 & 66.5 \\
            & \checkmark & \checkmark & \cellcolor{green!20}\textbf{84.8} & 59.0 & 69.6 \\
\hline
 \checkmark & \checkmark &            & 77.0 & \cellcolor{orange!20}20.9 & \cellcolor{orange!20}32.8 \\
 \checkmark & \checkmark & \checkmark & 79.7 & \cellcolor{green!20}\textbf{93.0} & \cellcolor{green!20}\textbf{85.8} \\
\hline
\end{tabular}}
\caption{Results of ablation on our QA benchmark with significant values highlighted.}
\label{table:ablation}
\end{table}

\subsection{Relationship between Faithfulness and Hallucination}
\label{sec:exp:faithfulnesshallucination}

We also conduct the experiment of our approach within a traditional QA setting. To our knowledge, it is the first work to elucidate the relationship between the faithfulness to context and to parametric knowledge (hallucination). R-Tuning \cite{zhang2024rtuning} preconstructs the tuning datasets to explicitly convey uncertainty for unknown questions, while we directly tune our pre-trained model on raw samples, as detailed in Appx. \ref{sec:hallusetting}. According to Table \ref{table:comp:rtuning}, while our US-Tuning shows lower effectiveness compared to the SOTA approaches specifically designed for noncontextual QA tasks, it represents a significant improvement over the vanilla model, with increases of 11.30\%, 10.38\%, and 6.26\% in accuracy, respectively. Our findings indicate that our model can leverage uncertainty recognition as a metacapacity, effectively applying it in both contextual and noncontextual QA scenarios.

\begin{table}[H]
\centering
\resizebox{0.50\textwidth}{!}{
\begin{threeparttable}
\begin{tabular}{llccc}
\hline
Tuning    &      Model               &  ParaRel     & MMLU       & HaluEval \\
\hline
Vanilla        & Llama2             & 43.38        & 38.56          & 76.22 \\
\hline
\multirow{3}{*}{NC}      & Honesty\tnote{*}             & -            & 49.28      & \textbf{88.11}    \\
                                  & Calibration         & -            & 53.00      & 87.78     \\
                                  & R-Tuning            & \textbf{69.54}   & \textbf{55.56} & 77.17 \\
\hline
C     & \textbf{US-Tuning}  & 54.68   & 48.94     & 82.48 \\
\hline
\end{tabular}
\begin{tablenotes}
        \footnotesize
        \item * Based on Llama2-13B-Chat \cite{touvron2023llama}
      \end{tablenotes}
  \end{threeparttable}
}
\caption{
Average Precisions (\%) of SOTA methods designed for noncontextual (NC) and contextual (C) QA tasks on QA hallucination detection benchmarks.
}
\label{table:comp:rtuning}
\end{table}

\begin{table}[H]
\centering
\resizebox{0.50\textwidth}{!}{
\begin{tabular}{lc||lcc}
\hline
Model             &       Vanilla    &    Model             &       Vanilla     &       \textbf{US-Tuning}  \\
\hline
 GPT-4           & 92.05            &  Llama2              & 94.04             & \textbf{94.37}  \\
 GPT-3.5          & 77.81            & Mistral             & 96.36             & \textbf{96.70}  \\
 Vicuna           & 82.62            & Gemma                & \textbf{95.28}    & 95.04   \\
\hline
\end{tabular}
}
\caption{
Compliance rate (\%) of prevalent LLMs on the CoCoNot noncontextual unknown QA benchmark.
}
\label{table:coconot}
\end{table}

Furthermore, CoCoNot \cite{brahman2024art} provides 302 unknown noncontextual QA pairs and suggests employing GPT-3.5 \cite{gpt3.5} to assess compliance. We test our pre-trained models on a subset of CoCoNot, and our results indicate that US-Tuning can also slightly improve the performance in rejecting noncontextual questions.

\section{Conclusion}

This paper investigates a prevalent issue in large language models (LLMs), where insufficient contextual information results in plausible yet incorrect responses. Our research reveals that LLMs often struggle with unknown questions, primarily due to their limited uncertainty recognition capacity and weak robustness to zero-shot instructions. Notably, tuning the models to focus on uncertainty will adversely weaken adherence to zero-shot instructions. To address these issues, we propose a novel two-stage training framework, termed "uncertainty-and-sensitive-aware tuning." The first stage guides the LLM to identify unknown questions, while the second stage aims to recover diminished question-answering performance through carefully designed causal instructions. This approach enhances the model's reliability and reduces hallucinations. Our methodology distinguishes itself by fine-tuning the uncertainty recognition as a metacapacity, rather than direct training on unknown question samples, thereby enabling effective adaptation across various tasks. By open-sourcing this work, we aim to advance the development of automatic instruction synthesis datasets, emphasising data diversity and the critical reduction of hallucinations.

\section*{Limitations}

In this study, we identify two key areas for future refinement. First, the LLM encounters a long-tail problem when tuned with datasets that contain a limited number of unknown questions, necessitating further adaptation of our US-Tuning. Second, we have not analysed the parametric knowledge acquired by Llama2 during its pre-training phase, and our fine-tuning dataset may overlap with this pre-training data, potentially affecting performance. To address these challenges, future research will investigate methods for measuring model uncertainty through internal parameter monitoring, as proposed by \citet{lu2023prompts}. By quantifying uncertainty across various inputs, we aim to identify knowledge gaps and long-tail weaknesses, informing targeted fine-tuning strategies to enhance the LLM's performance across diverse queries.

\section*{Ethics Statement}
The benchmark and datasets utilised in this study are derived from public datasets. Additionally, the US-Tuning dataset incorporates refinements using GPT-4, which may introduce inherent biases. However, the methodologies in this research are designed to avoid introducing any additional biases beyond those already inherent in the datasets.

% Bibliography entries for the entire Anthology, followed by custom entries
%\bibliography{anthology,custom}
% Custom bibliography entries only
\section*{Acknowledgment}
This work is partially supported by a research grant provided by HSBC.

\bibliography{anthology,custom}

%\newpage
\clearpage
\appendix

\section{Supplementary Material}

\subsection{Algorithm for Instruction Review Module}

Here we provide the algorithm chart for the Review Instruction Synthesis in Section \ref{section:ps}.

\begin{algorithm}
\KwData{context $c$, query $q$, task instruction $i_t$, causal instructions $i_c$}

\While{not fulfilled}
{
answer = generate($c$, $q$, $i_t$, $i_c$)\;
check = review(answer, $i_t$, $i_c$)\;
\If{"<not fulfilled>" not in check}
    {fulfilled = True\;}
    {}
}
\caption{Instruction Review Module}
\label{algo:revise}
\end{algorithm}

\subsection{Postfix Uncertainty-Recognition}

In addition to the question-answering (QA) Uncertainty-Recognition Benchmark mentioned in Section \ref{sec:data_construction}, we further develop a postfix template specifically for uncertainty recognition. Different from the QA one, the postfix template emphasises the assessment of uncertainty by evaluating the sufficiency of the responses and generating a tag after the corresponding answer. The prompt template is recorded as follow:

\begin{itemize}
\item Postfix Uncertainty-Recognition: \emph{You must append either ’<Sufficient>’ or ’<Insufficient>’ after your answer.}
\end{itemize}

\begin{table}[H]
\centering
\resizebox{0.50\textwidth}{!}{\begin{tabular}{llccc}
\hline
& & \multicolumn{3}{c}{Postfix Uncertainty-Recognition}\\
Category                    & Model                & $Acc_{known}$ & $Acc_{unknown}$  & $F1$ \\
\hline
\multirow{4}{*}{Benchmark}  & \textbf{GPT-4} & 88.9 & \cellcolor{green!20}\textbf{78.3} & \cellcolor{green!20}\textbf{83.3} \\
                            & GPT-3.5 Turbo        & \cellcolor{green!20}\textbf{97.0} & 33.4 & 49.7 \\
                            & Vicuna-7B v1.5       & 93.5 & \cellcolor{orange!20}14.3 & \cellcolor{orange!20}24.8 \\
                            & Self-RAG-7B          & \cellcolor{orange!20}46.0 & \cellcolor{green!20}74.9 & 57.0 \\
\hline
\multirow{4}{*}{Llama2}     & Vanilla   & 85.2 & 29.5 & 43.9 \\
                            & UT (Stage 1)         & 81.3 & \cellcolor{green!20}\textbf{84.0} & \cellcolor{green!20}\textbf{82.6} \\
                            & UT+HotpotQA        & 87.1 & \cellcolor{green!20}66.7 & \cellcolor{green!20}75.5 \\
                            & \textbf{US-Tuning}   & \textbf{88.0} & \cellcolor{green!20}66.0 & \cellcolor{green!20}75.4\\
\hline
\multirow{4}{*}{Mistral}    & Vanilla   & 82.8 & 43.1 & 56.7\\
                            & UT (Stage 1)         & \textbf{86.1} & \cellcolor{green!20}81.9 & \cellcolor{green!20}\textbf{84.0}\\
                            & UT+HotpotQA        & 80.7 & \cellcolor{green!20}75.1 & \cellcolor{green!20}77.8 \\
                            & \textbf{US-Tuning}   & 82.5 & \cellcolor{green!20}\textbf{82.2} & \cellcolor{green!20}82.4 \\
\hline
\multirow{4}{*}{Gemma}      & Vanilla   & 86.3 & 57.6 & 69.1\\
                            & UT (Stage 1)         & 93.4 & \cellcolor{green!20}\textbf{76.2} & \cellcolor{green!20}\textbf{83.9} \\
                            & UT+HotpotQA        & \textbf{99.4} & 58.7 & 73.8 \\
                            & \textbf{US-Tuning}   & 96.1 & 55.1 & 70.1 \\
\hline
\end{tabular}}
\caption{Results (in \%) for prevalent LLMs on postfix uncertainty-recognition benchmark. The overall best results are highlighted in \textbf{bold}.
Results that are more than 5\% higher or lower than the baseline are highlighted in \colorbox{green!20}{green} and \colorbox{orange!20}{orange}, respectively.}
\label{table:binaryUR}
\end{table}

Figure \ref{table:binaryUR} presents the evaluation results from our benchmark using the postfix template, focusing solely on the accuracy of the sufficiency tags rather than the correctness of answers. The findings indicate that most prevalent large language models (LLMs) struggle to effectively identify uncertainty. Furthermore, our proposed uncertainty-aware tuning (UT) shows potential to mitigate this challenge.

\subsection{Illustration to the Benchmark Results}
\label{sec:results}

Table \ref{table:benchmark} presents the QA performance on our benchmark. Coupled with the uncertainty recognition performance detailed in Table \ref{table:binaryUR}, our findings indicate that prevalent LLMs face challenges in accurately identifying unknown questions, achieving only approximately 60\% accuracy. Notably, GPT-4 and Gemma-2 achieve higher accuracies of 83.6\% and 74.1\%, respectively. Mistral and Llama-2 rank highest among the remaining models, surpassing GPT-3.5 despite its larger parameter size. Nevertheless, a significant performance gap persists between GPT-4 and other models. Ongoing experiments aim to explore the underlying factors contributing to this disparity. The analysis further reveals that different models respond differently to insufficient queries. Models fine-tuned on dialogue tasks tend to overly rely on and trust the given information. Self-RAG, which is fine-tuned for QA tasks involving unknown questions, demonstrates a strong ability to identify uncertainty, as indicated in Table \ref{table:binaryUR}, but still struggles to acknowledge it.

\subsection{Illustration to the State-of-the-Art (SOTA) Methods}
\label{sec:sotamethods}

Current SOTA research primarily addresses rejection in noncontextual QA tasks, leaving contextual QA underexplored. We categorise SOTA methodologies into post-generation, prompt-based, and tuning methods. Notable tuning approaches for rejecting unknown questions include Honesty-Alignment \cite{yang2023alignment} and Calibration-Tuning \cite{kapoor2024calibration}. They focus on noncontextual QA tasks while tuning for rejecting answering in contextual QA tasks remains unaddressed. C-DPO \cite{bi2024context} emphasizes model faithfulness to context rather than rejecting unknown questions.  Context-Faithful-Prompting (CFP) \cite{zhou-etal-2023-context} aims to enhance model fidelity to context through third-person paraphrasing in prompts. Post-generation methods for uncertainty detection include Multi-Sampling \cite{cole-etal-2023-selectively} and LM-Validation \cite{kadavath2022language}. The sampling method generates three outputs at a temperature of 0.6, selecting the most frequent response, while LM-Validation allows for further refinement of the generation. This study compares these methodologies with our proposed US-Tuning.

\begin{table}[H]
\large
\centering
\resizebox{0.5\textwidth}{!}{\begin{tabular}{l|ccc}
\hline
Model               &     Category     &     Contexutal         &  Task       \\
\hline
Validation          &     Post-Gen.    &        Both            &  Faithfulness\\
Sampling            &     Post-Gen.    &        Both            &  Faithfulness\\
CFP                 &     Prompt       &        Contextual      &  Faithfulness\\
Calibration         &     Tuning       &        Noncontextual   &  Rejection \\
Honesty             &     Tuning       &        Noncontextual   &  Rejection\\
C-DPO               &     Tuning       &        Contextual      &  Faithfulness\\
\textbf{ US-Tuning} &     Tuning       &        Contextual      &  Rejection \\
\hline
\end{tabular}}
\caption{Categories and targeted tasks for the SOTAs.}
\label{table:sotacategory}
\end{table}

\subsection{Further Ablation Study on the SOTA Method with Sensitivity-Aware Tuning}
\label{sec:st_honesty}

In Section \ref{sec:exp:sotacomperision}, we evaluate the performance of the SOTA methods on our QA uncertainty-recognition benchmark. We attribute the low performance of Honesty-Alignment \cite{yang2023alignment} to the instruction-reduction problem, evidenced by an $Acc_{unknown}$ of only 61.1\%, despite it being tuned on unknown noncontextual QA samples. In contrast, our US-Tuned Llama2 achieves 93.0\%. This section further elucidates the instruction-sensitivity reduction problem by implementing our Sensitivity-Aware Tuning (ST), which aims to enhance the model's sensitivity to constraint instructions alongside the Honesty-Alignment approach.

\begin{table}[H]
\centering
\resizebox{0.46\textwidth}{!}{\begin{tabular}{lccc}
\hline
Method & $Acc_{known}$ & $Acc_{unknown}$ & $F1$ \\
\hline
Vanilla Llama2  & 79.3 & 58.3 & 67.2 \\
\textbf{US-Tuning} & \textbf{79.7}    & \textbf{93.0} & \textbf{85.8} \\
\hline
Honesty            & 74.7    & 61.1             & 67.2         \\
\textbf{Honesty + ST}       & \textbf{80.4}    & \textbf{80.8}             & \textbf{80.6}         \\

\hline
\end{tabular}}
\caption{Results of sensitivity-aware tuned Honesty-Alignment on QA uncertainty-recognition benchmark.}
\label{table:st_honesty}
\end{table}

Table \ref{table:st_honesty} yields several key conclusions. First, our proposed ST effectively mitigates the instruction-sensitivity reduction problem, improving the $Acc_{unknown}$ of Honesty-Alignment by 19.7\%, resulting in a 13.4\% enhancement in overall performance. Second, our initial stage, focused on assessing the sufficiency of the given context relative to the question, outperforms other methods, as demonstrated by a 5.2\% improvement of our US-Tuned Llama2 over the Sensitivity-Aware Tuned Honesty-Alignment.
This advancement is attributable to both the quality and quantity of the dataset used for ST, enabling the model to recognize knowledge gaps as a metacognitive capacity, as discussed in Section \ref{sec:exp:faithfulnesshallucination}. Finally, the samples utilized for Sensitivity-Aware Tuned Honesty-Alignment are strictly non-overlapping with our benchmark, further validating the effectiveness and robustness of our proposed US-Tuning approach.

\begin{table*}
\centering
\resizebox{\textwidth}{!}{\begin{tabular}{l}
\hline
\begin{tabular}[c]{@{}l@{}}
\textbf{Context:} This is a passage about Harry Potter:\\
An \textcolor{red}{epilogue "Nineteen Years Later" (set on} \textcolor{red}{1 September 2017)} describes the lives of the surviving characters\\ and the effects of Voldemort's death on the Wizarding World. In the epilogue, Harry and Ginny are married\\ with three children, and Ron and Hermione are married with two children.
\end{tabular} \\
\hline
\begin{tabular}[c]{@{}l@{}}
\textbf{Unknown Question:} What year were the \textcolor{red}{Harry} \textcolor{red}{Potter books first released}?
\end{tabular} \\
\hline
\begin{tabular}[c]{@{}l@{}}
\textbf{Vanilla Llama2:} Sure, I'd be happy to help! 
Based on the context you provided, the Harry Potter books were \\
first released \textcolor{red}{in the year} 
\textcolor{red}{1997}. The passage states that the epilogue is set on September1, 2017, which is \textcolor{red}{19}\\
\textcolor{red}{years after the events of the final book}
\textcolor{red}{ in the series, "Harry Potter and the Deathly Hallows," which was }\\
\textcolor{red}{published in 1997.} Therefore, the Harry Potter books were first released in 1997.
\end{tabular} \\
\hline
\begin{tabular}[c]{@{}l@{}}
\textbf{US-Tuned Llama2:} Based on the given context, 
the answer to the question "What year were the Harry Potter\\ books first released?" is:

\textcolor{red}{Not Provided.}\\

The context \textcolor{red}{does not provide sufficient}
\textcolor{red}{information to determine} the year the Harry Potter books were 
first\\ released. The passage  
\textcolor{red}{only mentions the epilogue set in 2017}, which is 19 years after the events of the books. \\Therefore,  the year of the books' release is \textcolor{red}{not mentioned}  \textcolor{red}{or implied in the given context}.
\end{tabular} \\
\hline
\end{tabular}}
\caption{An illustrative case from our proposed benchmark, along with the corresponding model responses before and after the implementation of US-Tuning.}
\label{table:casestudy}
\end{table*}

\subsection{Figure of Ablation Study}
\label{appx:ablation}

In Section \ref{sec:ablation}, we present a comparative analysis of each model configuration. Table \ref{table:ablationname} details the specific names associated with each setting.

\begin{table}[H]
\large
\centering
\resizebox{0.28\textwidth}{!}{\begin{tabular}{l|ccc}
\hline
% Name & \multicolumn{3}{c}{Components}\\
Model &     UT     &     HP     &     CI   \\
\hline
Llama2&            &            &            \\
\hline
 UT &\checkmark &            &            \\
HotpotQA &           & \checkmark &             \\
ST &         & \checkmark & \checkmark  \\
\hline
 UT \& HotpotQA &\checkmark & \checkmark &            \\
 US-Tuning &\checkmark & \checkmark & \checkmark \\
\hline
\end{tabular}}
\caption{Corresponding model name to each setting in the ablation study.}
\label{table:ablationname}
\end{table}

\begin{figure}[H]
  \centering  \includegraphics[width=0.41\textwidth]{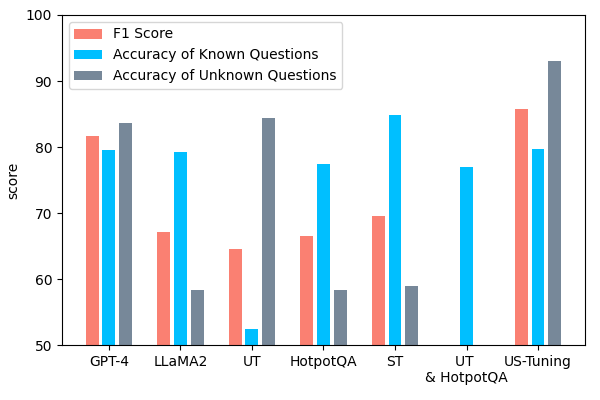}
  \caption{Comparison of different models in the ablation study. A higher score means better performance.}
  \label{fig:ablation}
\end{figure}

\subsection{Experiment Setting for Hallucination Benchmarks}
\label{sec:hallusetting}

R-Tuning \cite{zhang2024rtuning} divides the ParaRel \cite{elazar2021measuring} and MMLU \cite{hendrycks2020measuring} into in-domain and out-of-domain datasets, treating the entire HaluEval \cite{li2023halueval} as an out-of-domain evaluation set. Models are initially fine-tuned on in-domain datasets and subsequently evaluated on out-of-domain datasets. Our fine-tuning is based on Llama2-7B-Chat \cite{touvron2023llama} US-Tuned on ASQA \cite{stelmakh-etal-2022-asqa} and HotpotQA \cite{yang2018hotpotqa}, as detailed in Section \ref{sec:methodology}. We further fine-tune it on in-domain datasets with LoRA, adhering to the settings in R-Tuning: 1 epoch, a learning rate of 2e-5, and a batch size of 4. Similarly to R-Tuning, our evaluation spans several out-of-domain datasets.

CoCoNot \cite{brahman2024art} offers a comprehensive dataset for benchmarking the performance in rejecting answering. In this study, we utilise a subset containing 302 unknown QA pairs in a noncontextual setting and evaluate the effectiveness of US-Tuning on it. Following the methodology outlined in that paper, we employ GPT-3.5 \cite{gpt3.5} to assess the compliance with the response.

\subsection{Effectiveness on Noncontextual QA}
\label{sec:effectiveness_noncontextual}

Real-world applications include both contextual and noncontextual scenarios. We have conducted additional experiments to demonstrate that our two-stage tuning can be effectively applied in a noncontextual QA setting. We adhere strictly to the experimental setup of R-Tuning \cite{zhang2024rtuning} and use their pre-processed MMLU dataset, which is already divided into known and unknown questions. Specifically, 2,448 known questions are used for tuning, while the evaluation set consists of 2,439 known and 9,115 unknown questions.

First, we fine-tune the vanilla Llama2-7B-Chat on the 2,448 known questions. For UT, we further select 816 known and 816 unknown samples and ask the model with "Do you know the answer to X?". For ST, we use GPT-4o-mini \cite{gpt4o} to generate instruction adherence tuning data based on 300 ParaRel samples. The final model is evaluated on the remaining dataset, comprising 1,623 known and 8,299 unknown samples (The rest models in Table \ref{table:comp:noncontextual} are evaluated on the whole evaluation set).

Different from our benchmark, R-Tuning reports average precision, which involves both known and unknown questions. Here, in Table \ref{table:comp:noncontextual} , we provide the accuracies on known and unknown subsets of MMLU when setting the threshold to 0.5. 

Our noncontextual model outperforms the contextual one, demonstrating the generalisation and efficiency of our approach in real-world applications. Additionally, the performance variations can be attributed to potential data biases or inconsistencies in the training sets across different models, which is one reason why a contextual setting is preferred in our paper.

\begin{table}[H]
\centering
\resizebox{0.50\textwidth}{!}{
\begin{threeparttable}
\begin{tabular}{lcccc}
\hline
 Method & $Acc_{kno.}$ & $Acc_{unk.}$ & $F1$ & AP \\
\hline
Vanilla             & 57.13     & 34.37   & 42.92 & 38.56 \\
Honesty\tnote{*}            & 55.79     & 48.32   & 51.79 & 49.28 \\
Calibration         & 61.61     & 51.61   & 56.19 & 53.00 \\
R-Tuning            & 65.61     & 52.43   & 58.28 & 55.56 \\
\hline
US-Tuning (C)       & 56.07     & 45.63   & 50.31 & 48.94\\
\textbf{US-Tuning (NC)}      & 62.84     & 52.37   & 57.13 & 54.47\\
\hline
\end{tabular}
\begin{tablenotes}
        \footnotesize
        \item * Based on Llama2-13B-Chat \cite{touvron2023llama}
      \end{tablenotes}
  \end{threeparttable}
}
\caption{
Accuracies and average precisions (in \%) of SOTA methods on pre-processed MMLU benchmark.
Our method is fine-tuned under two configurations: C refers to the contextual setting, while NC denotes the non-contextual setting designed for this section.
}
\label{table:comp:noncontextual}
\end{table}

\subsection{Zero-Shot Effectiveness on RealtimeQA}
\label{sec:effectiveness_faithfulness}

RealtimeQA \cite{kasai2024realtime} is a dataset designed for high-stakes scenarios that necessitate timely responses, thereby challenging the faithfulness of LLMs to contextual information. Our study utilizes 113 contextual QA pairs from RealtimeQA, of which 50 are unknown pairs. Our benchmark is distinguished by a larger sample size compared to RealtimeQA. We directly implement our pre-trained model without further tuning on RealtimeQA.
As shown in Table \ref{table:effectiveness_faithfulness}, our model demonstrates significant improvements in addressing unknown questions, underscoring the effectiveness and robustness of our approach.

\begin{table}[H]
\centering
\resizebox{0.46\textwidth}{!}{\begin{tabular}{lccc}
\hline
Method & $Acc_{known}$ & $Acc_{unknown}$ & $F1$ \\
\hline
Vanilla Llama2  & \textbf{88.7} & 36.7 & 51.9 \\
\textbf{US-Tuning} & 71.8    & \textbf{56.0} & \textbf{62.9} \\

\hline
\end{tabular}}
\caption{Accuracies (\%) on RealtimeQA.}
\label{table:effectiveness_faithfulness}
\end{table}

\subsection{Case Study}
\label{sec:casestudy}

The case provided in Table \ref{table:casestudy} addresses a key challenge regarding uncertain information. The vanilla Llama2 incorrectly claims that the Harry Potter books were released in 1997, despite the context only referencing an epilogue set in 2017.

In contrast, our US-Tuned Llama2 effectively mitigates this issue by prioritising uncertainty detection. Rather than offering an uncertain answer, it appropriately responds with "Not Provided." This approach not only rejects uncertain responses but also clarifies the source of uncertainty, thereby enhancing the model's reliability. The implementation of US-Tuning is particularly vital in high-stakes fields, such as medicine, where a low wrong answer rate is essential. By refining LLMs' ability to recognise and communicate uncertainty, US-Tuning promotes responsible and trustworthy interactions, ensuring users receive reliable information.

\subsection{Example of Constructing Benchmark}
\label{sec:appendix_dataset}

\begin{table}[t]
\centering
\resizebox{0.5\textwidth}{!}{\begin{tabular}{l}
\hline
\begin{tabular}[c]{@{}l@{}}
    \textbf{Question:} Who discovered and developed an\\ explanationfor the photoelectric effect in \textcolor{red}{1887}?
\end{tabular} \\
\hline
\begin{tabular}[c]{@{}l@{}}
    \textbf{Positive Context:} This is a passage about\\ Photoelectric effect:
Light, and especially\\ ultra-violet light, discharges negatively electrified\\
bodies with the production of rays of the\\ samenature as cathode rays. 
Under certain\\ 
circumstances it candirectly ionize gases. The first\\ of
these phenomena was discovered by \textcolor{red}{Heinrich}\\
\textcolor{red}{Hertz and Wilhelm Hallwachs in 1887}.The second\\ was 
announced first by Philipp Lenard in 1900.
\end{tabular} \\
\hline
\begin{tabular}[c]{@{}l@{}}
    \textbf{Negative Context:} This is a passage about\\ Photoelectric effect:
\textcolor{red}{In 1905, Einstein} proposed\\ an explanation of the photoelectriceffect using a\\
concept first put forward by Max Planck that light\\ 
waves consist of tiny bundles or packets of energy\\
knownas photons or quanta. 
\end{tabular} \\
\hline
\end{tabular}}
\caption{An example from ASQA \cite{stelmakh-etal-2022-asqa}, where the positive context adequately supports the question, whereas the negative is insufficient.}
\label{table:benchmarkexample}
\end{table}

Table \ref{table:benchmarkexample} presents an example that illustrates the construction of our uncertainty-recognition benchmark, as detailed in Section \ref{sec:data_construction}. In this process, we shuffle the questions and their corresponding contexts to introduce uncertainty, thereby challenging the model's ability to respond to uncertain queries.

\section{Instructions}
\label{sec:appendix_instructions}

In this section, we present an overview of all the prompt templates utilized in this study. Key descriptions are highlighted in \textcolor{red}{red}, while \textcolor{blue}{blue} descriptions are designated for performance adjustments.

\subsection{Question Answering}
\label{sec:appendix_instructions:QA}

\textcolor{blue}{Question Answering Task: You need to do the Question Answering for the following query.}

I will give a question and several contexts.
Based on the given contexts, give an answer to the question.
Your answer must not using any additional knowledge that is not mentioned in the contexts.
If the context is not sufficient to answer the question, please answer it with 'Not Provided'

QUERY: $q$

CONTEXT: $c$

ANSWER:

\subsection{Uncertainty-Aware Tuning}
\label{sec:appendix_instructions:CA}

\textcolor{blue}{Cognition Assessment Task: You need to do the Cognition Assessment for the following query. }

I will give a query and a related context about the query.
Your task is to judge whether the context is sufficient to answer the query.

\textcolor{red}{Assessment: You must append either '<Sufficient>' or '<Insufficient>' after your answer.}

\textcolor{red}{Finetuning: You must only answer either 'Sufficient' or 'Insufficient' without any other output.}

Here is the example.

QUERY:
What happened to Jay when he got old?

CONTEXT:
Jay Chou was the most famous singer in China when he was young, releasing many nostalgic albums and songs that are memorable to middle-aged people today.

ANSWER:

\textcolor{red}{Assessment: Jay Chou was the most famous singer in China.<Insufficient>}

\textcolor{red}{Finetuning: Insufficient}

Here is the provided information that you need to accomplish follow the provided example:

QUERY: $q$

CONTEXT: $c$

ANSWER:

\subsection{Additional Casual Instruction Generation}
\label{sec:appendix_instructions:PSGeneration}

Your task is to provide various instructions for the questions answering task. 

The questions answering task provides a context and a query. e.g. "Context: XXX Query: XXX Answer:". 
And your task is to add some specific requirement to the answer. e.g. "The answer must be all in upper case", "There should be no punctuation in the answer". 
The added instruction should be general to the query.
You should generate hundreds of such instructions.

\subsection{Sensitivity-Aware Tuning}
\label{sec:appendix_instructions:PS}

You should check whether your answer aligned the requirement by generating a Checking part, 
\textcolor{red}{checking each sentence of the above instruction, with either <fulfilled> or <not fulfilled> mark behind the sentence, indicating whether the requirement is fulfilled or not}.
If there is <not fulfilled> mark behind the sentence, you must modify your answer again to fulfill the requirement, by appending a new ANSWER and CHECKING part.

Here is an example for this task:

e.g. Question Answering Task Requirements: You need to do the Task Prompt for the following query and context.
\textcolor{red}{Ensure the response is written in the past tense.}

QUESTION: Who is Jack Chen?

CONTEXTS: People saying that Jack Chen is a famous singer in China. 

ANSWER: Jack Chen is a famous singer in China.

CHECKING: Question Answering Task: You need to do the Task Prompt for the following query and context.<fulfilled>Ensure the response is written in the past tense.<not fulfilled>

ANSWER: Jack Chen is a famous singer in China.

CHECKING: Question Answering Task: You need to do the Task Prompt for the following query and context.<fulfilled>Ensure the response is written in the past tense.<fulfilled>

Here is the information of your task:

\{Question Answering Instruction\}

\subsection{Trustworthy Question Answering for Benchmark}
\label{sec:appendix_instructions:TQA}

\textcolor{blue}{Trustworthy Question Answering Task:  You need to utilize the ability learnt during both the Question Answering Task and Cognition Assessment Task. And only provide the answers which are sufficiently supported by the context, otherwise provide 'Not Provided'}

I will give a question and several context texts about the question.
Based on the given contexts, give an answer to the question.
Your answer must not using any additional knowledge that is not mentioned in the given contexts.
\textcolor{red}{If the context is not sufficient to answer the question, please answer it with 'Not Provided'}

QUERY: $q$

CONTEXT: $c$

ANSWER:

\section{Causal Instructions}
\label{appx:causal_ins}

We generated 100 causal instructions using GPT-4, as detailed in the prompts recorded in Appx. \ref{sec:appendix_instructions:PSGeneration}. Subsequently, we manually selected the 28 most effective instructions based on criteria of robustness. For instance, "Answer in chronological order" is deemed lacking in robustness, as many responses do not conform to a chronological structure. Followinges the causal instructions we employed:

\noindent
1. Ensure the answer is summarised in less than 50 characters.

\noindent
2. Include at least three potential answers in the response.

\noindent
3. Include examples from the context.

\noindent
4. Express the answer using bullet points.

\noindent
5. Limit the response to a minimum of 20 words.

\noindent
6. Ensure the response is written in the past tense.

\noindent
7. Provide a concise definition of each answer.

\noindent
8. Provide a wrong answer that did occurr in the context but not the answer to the query.

\noindent
9. Present the answer as a dialogue between two characters discussing the topic.

\noindent
10. Incorporate elements of humour or wit into the response.

\noindent
11. Provide the answer in a complete sentence.

\noindent
12. Provide a brief explanation using terminology.

\noindent
13. Include a relevant metaphor or analogy to explain the concept

\noindent
14. Incorporate a fictional example or event into it.

\noindent
15. Frame the answer as a hypothetical scenario or speculation.

\noindent
16. Write the answer in the style of a news headline or tabloid headline.

\noindent
17. Frame the answer as a philosophical reflection on the question.

\noindent
18. Present the answer as a list of humorous alternatives or alternatives.

\noindent
19. Use creative storytelling techniques to answer.

\noindent
20. Include a riddle or puzzle that indirectly hints at the answer.

\noindent
21. Write in the style of a poem or lyrics.

\noindent
22. Include a fictional quote or excerpt from a fictional text that relates to the topic.

\noindent
23. Use imagery or descriptive language to paint a vivid picture of the answer.

\noindent
24. Write the answer in the form of a limerick or tongue twister.

\noindent
25. Incorporate elements of suspense or mystery into the response.

\noindent
26. Use hyperbole or exaggeration to emphasise a point in the response.

\noindent
27. Incorporate elements of fantasy or science fiction into the response.

\noindent
28. Use symbolism or allegory to convey deeper meaning in the response.

\end{document}